\documentclass[acmlarge,nonacm]{acmart}

\AtBeginDocument{%
  }

\usepackage{multirow,array}
\usepackage{listings}
\usepackage{adjustbox}
\usepackage{graphicx}
\usepackage{tabularx}

\begin{document}

\title{Leveraging Large Language Models for Identifying Knowledge Components}

\author{Canwen Wang}
\email{canwenw@andrew.cmu.edu}
\authornotemark[1]
\affiliation{%
  \institution{Carnegie Mellon University}
  \city{Pittsburgh}
  \state{Pennsylvania}
  \country{USA}
}

\author{Jionghao Lin}
\affiliation{%
  \institution{Carnegie Mellon University}
  \city{Pittsburgh}
  \country{USA}}
\email{jionghal@andrew.cmu.edu}

\author{Kenneth R. Koedinger}
\affiliation{%
  \institution{Carnegie Mellon University}
  \city{Pittsburgh}
  \country{USA}
}








\settopmatter{printacmref=false}
\maketitle

\section{Introduction}
Knowledge Components (KCs), defined as ``an acquired unit of cognitive function or structure that can be inferred from performance on a set of related tasks'',
refer to various cognitive elements---from technical constructs like production rules and schema to everyday concepts such as principles, facts, and skills \cite{koedinger2012knowledge}. KCs form the bedrock of skill models in intelligent educational software \cite{nguyen2019using}, playing a pivotal role in adaptive learning systems that rely on thousands of practice items organized into hundreds of components within a domain model \cite{pelanek2020managing}.  


While KCs provide significant advantages for instructional design and adaptive learning, 
their identification and extraction have traditionally depended on human effort through methods like Cognitive Task Analysis \cite{clark2008cognitive}. This manual process requires domain experts to label and extract KCs, and is highly labor-intensive~\cite{cooke1994varieties,shi2024knowledge}. 
Large language models (LLMs) have recently been used for automatic knowledge component extraction, as introduced by Moore et al.~\cite{moore2024automated}. They proposed two prompting strategies for generating KCs in multiple-choice questions (MCQs): simulated expert and simulated textbook, but they evaluated both approaches on a small set of 80 MCQs; furthermore, as acknowledged by the authors, both approaches produced superfluous KC labels with slightly different wording for questions that a teacher would label with a single KC. 

Addressing the above-mentioned limitations, this study builds on the simulated textbook approach to extract KCs from the E-Learning dataset\footnote{\url{https://pslcdatashop.web.cmu.edu/DatasetInfo?datasetId=5426}} 
with two key contributions: 

\begin{enumerate}
    \item We scaled the method to larger datasets---from 80 to 646 MCQs---and assessed its performance.  
    \item We calculated the cosine similarity between KC labels and merged similar ones at three similarity thresholds to remove redundant labels. 
\end{enumerate}




\section{Methods}

We adopted similar KC extraction steps as outlined in \cite{moore2024automated}, leveraging a simulated textbook strategy and utilizing the OpenAI API \cite{auger2024overview} for automatic generation of KCs. Through a multi-turn conversation with the GPT-4o-mini model (\texttt{gpt-4o-mini-2024-07-18}), our KC extraction pipeline consists of three main steps depicted in Table~\ref{prompt}.

During the first turn, we ask GPT to identify five domain-specific topics for an MCQ using the prompt under \texttt{[First Turn]}; we then ask it to  paraphrase the identified topics with verbs reflecting Bloom’s Revised Taxonomy \cite{krathwohl2002revision}  (\texttt{[Second Turn]}), and finally, to choose the best topic as the KC for the given MCQ (\texttt{[Last Turn]}). The 646 MCQs used in our analysis come from the E-learning dataset hosted on DataShop\footnote{\url{http://learnlab.org/datashop}} \cite{koedinger2010data} with Dataset ID 5426.




To evaluate our results, we compare the GPT-generated KCs with an expert-designed KC model \texttt{LOs-new-MCQ}, which consists of teacher-coded learning objectives, through a three-fold cross-validation
where we fit an Additive Factor Model (AFM)~\cite{afm} using both KC models. We ran cross-validation ten times with different random seeds and used the average item-blocked Root Mean Squared Error (RMSE) \cite{stamper2011human} as the evaluation metric. To compute item-block RMSE, we evenly distributed student steps (i.e., items) into the three cross-validation folds, following the standard practice of DataShop; a lower RMSE indicates a better KC model. 

\begin{lstlisting}[frame=single,basicstyle=\ttfamily,showspaces=false,caption=The prompts we adapted from~\cite{moore2024automated} to extract KCs through a three-turn conversation with GPT-4o-mini,captionpos=t,escapechar=\%,label=prompt]
%{\textbf{[System Prompt]}%
Below there is a multiple-choice question intended for a master's level 
audience with existing prior knowledge on the subject of E-learning. 
The question is used as a low-stakes assessment as part of a master's level 
E-learning course that covers similar content.

%\textbf{[First Turn]}%
If this question was presented in a textbook for a master's level 
E-learning course, what five domain-specific low-level detailed topics 
would the page cover? Note that the question is for a college audience 
with existing prior knowledge in E-learning.
Question text:
{stem}
{choices}

%\textbf{[Second Turn]}%
{topics identified by GPT}
Based on these topics, reword them to begin with action words from 
Bloom's Revised Taxonomy, while keeping them domain-specific, low-level, 
and detailed.

%\textbf{[Last Turn]}%
{topics reworded by GPT}
Of these topics, which is the most relevant to the question?

\end{lstlisting}

\section{Results and Discussion}

\begin{figure}
    \centering
    \includegraphics[width=\textwidth]{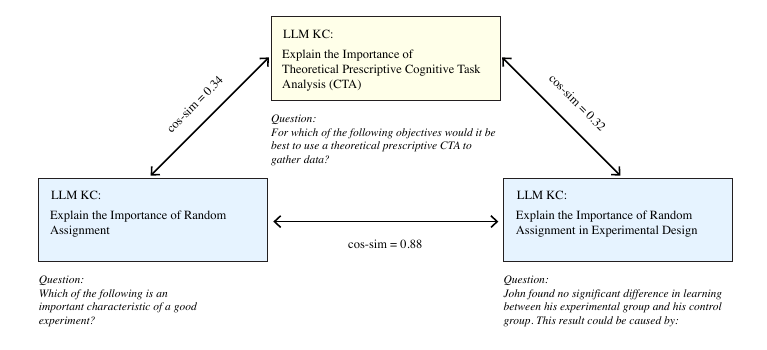}
    \caption{An example illustrating the extracted KC labels and their cosine similarity between three MCQs}
    \label{fig: cosine}
\end{figure}

In our study, we employ a three-prompt strategy to extract KCs;
the resulting model, which identified 569 KCs from 646 MCQs, achieved a RMSE of 0.4285---a value significantly ($t(18) = 12.8944, p < .001)$ higher than that of expert-designed \texttt{LOs-new-MCQ} model, which has 101 KCs and a RMSE of 0.4206. The simulated textbook approach proposed by~\cite{moore2024automated}, when scaled to more MCQs, did not perform better than the expert-designed model, although GPT-4o-mini is equally capable (if not more) as GPT-4 used in~\cite{moore2024automated}.


As acknowledged by the authors~\cite{moore2024automated}, a limitation of the simulated textbook approach is that it produces an excessive number of KCs, some of which correspond to one expert-designed KC but have slightly different wording. Figure~\ref{fig: cosine} shows an example where the bottom two MCQs are both about random assignment but GPT labeled one MCQ with the extra phrase ``in experimental design'' that adds no new information, causing the two MCQs to be labeled with separate KCs. 


To eliminate potentially superfluous KCs, we used OpenAI text embedding API to generate embeddings for the extracted KC labels; we then calculated cosine similarity, a widely used metric in text mining and natural language processing \cite{li2013distance}, to measure the text similarity between GPT-generated KC labels. Our strategy is to merge KCs whose labels have a cosine similarity above a predefined threshold (e.g., 0.8), thereby consolidating KCs that have
semantically close labels into one KC. As illustrated in Figure~\ref{fig: cosine}, the semantically similar KCs of the bottom two MCQs have a much higher cosine similarity of 0.88 compared to a distinct KC shown on the top, although all KCs begin with ``Explain the importance of...''. If we use 0.8 as a threshold, the bottom two KCs would be merged, with the top KC left intact.



Since a cosine value of 1 indicates an exact text match, we experimented with three thresholds close to 1 to merge only near matches: 0.9, 0.8, and 0.7. As shown in Table~\ref{tab: threshold}, the resulting models, using a threshold of 0.9, 0.8, and 0.7, produced 511, 428, and 273 KCs with RMSE values of 0.4264, 0.4259, and 0.4270, respectively. The best model, obtained at a cosine threshold of 0.8, significantly improves the RMSE from 0.4285 in the vanilla simulated textbook approach to 0.4259 ($t(18) = 5.1306, p < .001$), indicating a smaller discrepancy with the expert-designed model \texttt{LOs-new-MCQ}. Our approach to merge similar KCs by cosine similarity not only resulted in a simpler KC model with fewer KCs but a better model as measured by RMSE.



\section{Conclusion}

In this work, we extended the simulated textbook approach~\cite{moore2024automated} to extract KCs using LLM to a larger dataset of 646 MCQs and showed that this approach not only performed significantly worse than the expert-designed KC model but also produced extraneous KC labels with slightly different wording for questions that a teacher would label with a single KC. To alleviate the limitations of the original approach, we proposed merging semantically similar KCs based on cosine similarity and demonstrated, via three-fold cross-validation, that it
improved prediction over the previous automated method. In future work, we will continue refining the prompts to reduce the performance gap we observed between LLM-generated and expert-designed KC models, and use our refined KC model to reveal insights into those expert-designed KCs that show no learning~\cite{koedinger2012automated}, to help improve instruction. 

\setcounter{table}{1}
\begin{table}[t]
    \centering
        \caption{Summary of KC counts and RMSE across three thresholds}
    \label{tab: threshold}
    \resizebox{0.45\textwidth}{!}{
    \begin{tabular}{ccc}
    \toprule
       Cosine Threshold  & KC Counts & Item-blocked RMSE \\
       \midrule
        $\geq$ 0.9 & 511 & 0.4264\\
        $\geq$ 0.8 & 428 & 0.4259 \\
        $\geq$ 0.7 & 273 & 0.4270 \\
    \midrule
    \texttt{LOs-new-MCQ} & 101 & 0.4206 \\
    \bottomrule
    \end{tabular}
    }

\end{table}

\begin{acks}
Special thanks to Yumou Wei for providing the 646 MCQs and helping revise the first draft of this paper. 
\end{acks}

\bibliographystyle{ACM-Reference-Format}
\bibliography{Test_references}










\end{document}